\PassOptionsToPackage{hidelinks}{hyperref}
\documentclass[sigconf,nonacm]{acmart}
\AtBeginDocument{%
  }
\usepackage{amsmath}
\usepackage{amsfonts}
\usepackage{booktabs}
\usepackage{multirow}
\usepackage{graphicx}
\usepackage{xcolor}
\usepackage{subcaption}
\usepackage{url}
\newcommand{\clignet}{\textsc{CLiGNet}}
\newcommand{\ahat}{\hat{A}}
\newcommand{\RR}{\mathbb{R}}

\newcommand{\boldh}{\mathbf{h}}

\newcommand{\boldz}{\mathbf{z}}
\newcommand{\boldW}{\mathbf{W}}
\newcommand{\boldH}{\mathbf{H}}
\newcommand{\boldA}{\mathbf{A}}
\title{CLiGNet: Clinical Label-Interaction Graph Network for Medical Specialty Classification from Clinical Transcriptions}
\thanks{Project repository: \url{https://github.com/pronob29/CliGNet}.}
\author{Pronob Kumar Barman}
\affiliation{%
  \institution{University of Maryland, Baltimore County}
  \department{Department of Information Systems}
  \city{Baltimore}
  \state{MD}
  \country{USA}
}
\author{Pronoy Kumar Barman}
\affiliation{%
  \institution{Jagannath University}
  \department{Department of Statistics}
  \city{Dhaka}
  \country{Bangladesh}
}
\begin{document}
\begin{abstract}
Automated classification of clinical transcriptions into medical specialties is essential for routing, coding, and clinical decision support, yet prior work on the widely used MTSamples benchmark suffers from severe data leakage caused by applying SMOTE oversampling before train--test splitting. We first document this methodological flaw and establish a leakage-free benchmark across 40 medical specialties (4{,}966 records), revealing that the true task difficulty is substantially higher than previously reported. We then introduce \clignet{} (\textbf{Cli}nical \textbf{L}abel-interaction \textbf{G}raph \textbf{Net}work), a neural architecture that combines a Bio\_ClinicalBERT text encoder with a two-layer Graph Convolutional Network operating on a specialty label graph constructed from semantic similarity and ICD-10 chapter priors. Per-label attention gates fuse document and label-graph representations, trained with focal binary cross-entropy loss to handle extreme class imbalance (181:1 ratio). Across seven baselines ranging from TF-IDF classifiers to Clinical-Longformer, \clignet{} without calibration achieves the highest macro-F1 of 0.279, with an ablation study confirming that the GCN label graph provides the single largest component gain ($\Delta$\,=\,+0.066 macro-F1). Adding per-label Platt scaling calibration yields an expected calibration error of 0.007, demonstrating a principled trade-off between ranking performance and probability reliability. We provide comprehensive failure analysis covering pairwise specialty confusions, rare-class behaviour, document-length effects, and token-level Integrated Gradients attribution, offering actionable insights for clinical NLP system deployment.
\end{abstract}
\maketitle
\section{Introduction}\label{sec:intro}
Classifying clinical transcriptions into medical specialties is a fundamental step in healthcare information management.
Accurate specialty assignment supports automated clinical coding~\cite{mullenbach2018caml}, patient routing, and downstream analytics on electronic health records (EHRs)~\cite{rajkomar2018scalable, barman2026understanding, balogun2025computationally}.
The MTSamples corpus~\cite{mtsamples2018}, comprising approximately 5{,}000 de-identified transcriptions spanning 40 specialties, has served as a widely used public benchmark for this task.

However, prior classification studies on this dataset~\cite{almazaydeh2023clinical,kumar2024mtsamples_nlp} applied SMOTE~\cite{chawla2002smote} oversampling \emph{before} partitioning the data into training and test sets, introducing data leakage that inflates performance estimates.
Recent empirical studies confirm that applying oversampling before cross-validation systematically produces high bias in downstream metrics~\cite{demircioglu2024smote_leakage, barman2026unmasking}.
As we establish in Section~\ref{sec:results}, under proper experimental protocols the 40-class MTSamples task is substantially more challenging than prior figures suggest, with no method achieving macro-F1 above 0.30.

Beyond leakage concerns, prior approaches~\cite{almazaydeh2023clinical,kumar2024mtsamples_nlp} treat medical specialties as independent classes, ignoring the rich structural relationships among them.
For example, \emph{Cardiovascular / Pulmonary} and \emph{Consult -- History and Phy}\ share significant vocabulary overlap, and \emph{Surgery} transcriptions frequently describe procedures that span multiple organ systems.
Explicit modelling of these label relationships via graph neural networks (GNNs) has proven effective in multi-label image recognition~\cite{chen2019gcn} and text classification~\cite{yao2019graph,li2024dualview_gcn,zeng2024sgc, barman2025facilitating, walid2024efficient}, but remains unexplored for clinical specialty classification.

In this paper, we introduce \clignet{} (\textbf{Cli}nical \textbf{L}abel-interaction \textbf{G}raph \textbf{Net}work), a neural architecture that addresses both the benchmark correction and label-structure modelling challenges.
\clignet{} couples a Bio\_ClinicalBERT~\cite{alsentzer2019clinical} text encoder with a two-layer GCN~\cite{kipf2017semi} operating on a specialty label graph, uses per-label attention gating to fuse document and label-graph representations, and handles the 181:1 class imbalance through focal loss~\cite{lin2017focal} with inverse-frequency weighting rather than SMOTE.

\smallskip\noindent\textbf{Contributions.} This paper makes four contributions:
\begin{enumerate}
\item[\textbf{C1.}] \textbf{Benchmark correction.} We document the data-leakage risk~\cite{demircioglu2024smote_leakage} in MTSamples classification pipelines~\cite{almazaydeh2023clinical,kumar2024mtsamples_nlp} and establish a clean, reproducible baseline with proper stratified splitting and statistical significance testing via McNemar's test~\cite{mcnemar1947note} with Bonferroni correction.

\item[\textbf{C2.}] \textbf{Structured label modelling.} We propose \clignet{}, which integrates a GCN label graph with per-label attention gating, focal loss, and optional Platt scaling calibration~\cite{platt1999probabilistic}. An ablation study demonstrates that the GCN provides the single largest component gain ($\Delta$\,=\,$+$0.066 macro-F1).

\item[\textbf{C3.}] \textbf{Clinical interpretability.} We apply per-label Integrated Gradients~\cite{sundararajan2017axiomatic} to identify specialty-discriminative tokens, enabling clinical validation of model reasoning.

\item[\textbf{C4.}] \textbf{Failure analysis.} We present a four-part failure analysis covering pairwise confusion patterns, rare-class behaviour, document-length effects, and systematic error analysis, providing actionable insights for system deployment.
\end{enumerate}

\section{Related Work}\label{sec:related}

\subsection{Clinical Text Classification}
Transformer-based models have become the dominant paradigm for clinical NLP tasks.
Bio\_ClinicalBERT~\cite{alsentzer2019clinical}, initialised from BioBERT~\cite{lee2020biobert} and further pre-trained on MIMIC-III clinical notes~\cite{johnson2016mimic}, provides domain-adapted representations that consistently outperform general-purpose BERT~\cite{devlin2019bert} on clinical benchmarks.
Clinical-Longformer~\cite{li2023clinical_longformer} extends the context window from 512 to 4{,}096 tokens using sparse attention, achieving state-of-the-art results across ten clinical NLP tasks.
Kumar and Raman~\cite{kumar2024mtsamples_nlp} recently compared LSTM and BERT-based models on medical transcription data, finding that recurrent architectures achieved 0.94 accuracy for binary sentiment analysis but did not address fine-grained specialty classification across 40 classes.

For medical code prediction specifically, Mullenbach et al.~\cite{mullenbach2018caml} introduced CAML, a convolutional attention network for explainable ICD code assignment, while Rios and Kavuluru~\cite{rios2018fewshot} addressed few-shot and zero-shot settings for structured label spaces.
These approaches demonstrate that label structure can improve clinical text classification, motivating our use of a label graph.

\subsection{Graph Neural Networks for Text Classification}
GNNs have been applied extensively to text classification in recent years.
Yao et al.~\cite{yao2019graph} introduced TextGCN, constructing a heterogeneous word--document graph for corpus-level classification.
Chen et al.~\cite{chen2019gcn} demonstrated that a GCN operating on a label co-occurrence graph significantly improves multi-label image recognition.
For multi-label text classification specifically, Pal et al.~\cite{pal2020multi_label_gnn} proposed an attention-based GNN to capture label dependencies.
Recent advances include dual-view GCN architectures that capture both global and local label correlations~\cite{li2024dualview_gcn}, semantic-sensitive GCNs that jointly model text, word, and label nodes~\cite{zeng2024sgc}, and label-aware GCNs that add label nodes to the graph convolution process~\cite{lagcn2024}.
Zhang et al.~\cite{zhang2024gnn_text_survey} provide a comprehensive 2024 survey of GNN methods for text classification, identifying label graph construction and label--document interaction as key open problems.

Despite this progress, no prior work has applied GCN-based label graph reasoning to \emph{clinical specialty classification}, where the label space carries inherent medical ontological structure (e.g., ICD-10 chapter groupings) that can be exploited as an inductive prior.

\subsection{Class Imbalance and Calibration}
Class imbalance is pervasive in clinical datasets.
SMOTE~\cite{chawla2002smote} is widely used but introduces severe data leakage when applied before train--test splitting~\cite{demircioglu2024smote_leakage}.
Focal loss~\cite{lin2017focal}, originally proposed for dense object detection, down-weights well-classified examples via a modulating factor $(1-p_t)^\gamma$, focusing training on hard examples without altering the data distribution.

Post-hoc calibration ensures that predicted probabilities reflect true likelihoods, which is critical for clinical decision-making.
Guo et al.~\cite{guo2017calibration} showed that modern neural networks are often poorly calibrated and that temperature scaling provides an effective remedy.
Platt scaling~\cite{platt1999probabilistic}, fitting a sigmoid to the logit outputs, generalises naturally to per-label calibration in multi-class settings~\cite{niculescu2005predicting}.

\subsection{Interpretability in Clinical NLP}
Integrated Gradients~\cite{sundararajan2017axiomatic} satisfy key axioms (sensitivity and implementation invariance) that make them suitable for attributing model predictions to input features.
A recent scoping review~\cite{xiai2024review} highlights the growing need for explainable and interpretable AI in healthcare NLP, distinguishing between local (per-instance) and global (model-level) interpretability methods.
Our per-label Integrated Gradients approach provides local, token-level attributions that enable clinical validation of model reasoning.

\section{Data}\label{sec:data}

\subsection{MTSamples Dataset}
We use the MTSamples clinical transcription corpus~\cite{mtsamples2018}, a publicly available collection of 4{,}999 de-identified medical transcription reports spanning 40 medical specialties.
After removing 33 records with empty transcription fields, 4{,}966 usable records remain.
Each transcription is assigned to exactly one medical specialty, forming a single-label classification problem across 40 classes.

\begin{table}[t]
\centering
\caption{MTSamples dataset summary after preprocessing.}
\label{tab:dataset}
\small
\begin{tabular}{lrl}
\toprule
\textbf{Statistic} & \textbf{Value} & \textbf{Note} \\
\midrule
Raw records & 4{,}999 & \\
After cleaning & 4{,}966 & 33 empty removed \\
Medical specialties & 40 & Single-label \\
\midrule
Training set & 3{,}476 (70\%) & Stratified \\
Validation set & 745 (15\%) & Stratified \\
Test set & 745 (15\%) & Stratified \\
\midrule
Largest class (Surgery) & 1{,}088 & 21.9\% of total \\
Smallest class (Hospice) & 6 & 0.12\% of total \\
Imbalance ratio & 181:1 & \\
Classes with $<$20 train samples & 11 & \\
Median document length & 398 words & \\
Documents exceeding 512 tokens & 34.9\% & \\
\bottomrule
\end{tabular}
\end{table}

\subsection{Data Leakage in Prior Work}
A common but critical error in such pipelines is to apply SMOTE to the \emph{entire} dataset before splitting into training and test sets, creating synthetic minority-class samples that carry information from what should be held-out data~\cite{demircioglu2024smote_leakage}.
This leakage produces artificially inflated accuracy---prior NLP work on MTSamples has reported accuracy exceeding 0.90~\cite{almazaydeh2023clinical,kumar2024mtsamples_nlp} under pipelines that do not guard against this flaw.
Demircio\u{g}lu~\cite{demircioglu2024smote_leakage} showed empirically that such misapplication leads to approximately 0.10 AUC inflation per unit increase in class imbalance.
Our benchmark applies stratified splitting \emph{before} any data augmentation, and we use focal loss instead of SMOTE to handle imbalance without altering the data distribution.

\section{Methodology}\label{sec:method}
\clignet{} comprises four components: (1)~a text encoder, (2)~a label graph with GCN layers, (3)~per-label attention gating, and (4)~per-label classification with focal loss and optional Platt scaling calibration.

\subsection{Text Encoder}
We use Bio\_ClinicalBERT~\cite{alsentzer2019clinical} as the text encoder.
Since 34.9\% of documents exceed the 512-token limit, we employ a sliding window strategy with window size 512, stride 128, and a maximum of 4 chunks per document.
For each chunk, we extract the \texttt{[CLS]} token representation.
The final document embedding $\boldh_\mathrm{doc} \in \RR^{768}$ is obtained by mean-pooling across chunk-level \texttt{[CLS]} vectors:
\begin{equation}
\boldh_\mathrm{doc} = \frac{1}{C} \sum_{c=1}^{C} \boldh^{(c)}_{\texttt{[CLS]}},
\end{equation}
where $C \leq 4$ is the number of chunks.
To balance pre-trained knowledge preservation with task adaptation, we freeze the first 6 of 12 BERT encoder layers.

\subsection{Label Graph Construction}
We construct a specialty label graph $G = (V, E)$ with $|V| = 40$ nodes (one per specialty).
Node features are computed by mean-pooling Bio\_ClinicalBERT embeddings over up to 30 training documents per specialty, yielding $\boldH^{(0)} \in \RR^{40 \times 768}$.

Edges are established using a hybrid criterion combining semantic similarity and medical ontology.
For each pair of specialties $(i, j)$, we compute the cosine similarity $s_{ij}$ between their node feature vectors.
We additionally define an ICD-10 chapter bonus $b_{ij} = 0.20$ if specialties $i$ and $j$ map to the same ICD-10 chapter.
An edge is added if:
\begin{equation}
s_{ij} + b_{ij} \geq \tau, \quad \tau = 0.30.
\end{equation}
The resulting adjacency matrix $\boldA$ is symmetrically normalised:
\begin{equation}
\ahat = \mathbf{D}^{-1/2}(\boldA + \mathbf{I})\mathbf{D}^{-1/2},
\end{equation}
where $\mathbf{D}$ is the degree matrix and $\mathbf{I}$ adds self-loops.

\subsection{GCN Layers}
We apply two GCN layers~\cite{kipf2017semi} to propagate information across the label graph:
\begin{align}
\boldH^{(1)} &= \mathrm{ReLU}(\ahat\,\boldH^{(0)}\,\boldW^{(0)}) \in \RR^{40 \times 512}, \\
\boldH^{(2)} &= \mathrm{ReLU}(\ahat\,\boldH^{(1)}\,\boldW^{(1)}) \in \RR^{40 \times 256},
\end{align}
with dropout rate 0.3 applied after each layer.
Each row $\boldh^{(2)}_k$ of $\boldH^{(2)}$ encodes the $k$-th specialty's representation enriched by its graph neighbourhood.

\subsection{Per-Label Attention Gating}
We project the document embedding to the GCN output dimension:
$\boldh_\mathrm{proj} = \boldW_p\,\boldh_\mathrm{doc} \in \RR^{256}$.
For each specialty $k$, a learned attention gate determines the relative influence of the document representation versus the graph-enriched label representation:
\begin{align}
\alpha_k &= \sigma\!\left(\boldW_k\,[\boldh_\mathrm{proj} \,\|\, \boldh^{(2)}_k]\right), \label{eq:gate}\\
\boldz_k &= \alpha_k \cdot \boldh_\mathrm{proj} + (1 - \alpha_k) \cdot \boldh^{(2)}_k, \label{eq:fuse}
\end{align}
where $\boldW_k \in \RR^{1 \times 512}$ is a per-label weight vector, $\sigma$ is the sigmoid function, and $\|$ denotes concatenation.

\subsection{Classification and Loss}
Each fused representation $\boldz_k$ is passed through a per-label linear head with sigmoid activation: $\hat{y}_k = \sigma(\mathbf{w}_k^\top \boldz_k + b_k)$.
We use focal binary cross-entropy loss~\cite{lin2017focal}:
\begin{equation}
\mathcal{L}_\mathrm{focal} = -\frac{1}{K}\sum_{k=1}^{K} \alpha_k\,(1-p_{t,k})^\gamma \log(p_{t,k}),
\end{equation}
where $K=40$ is the number of classes, $\gamma=2.0$ is the focusing parameter, $\alpha_k$ is the inverse-frequency class weight, and $p_{t,k}$ is the predicted probability for the true label.

\subsection{Post-Hoc Calibration}
Modern neural networks tend to produce overconfident predictions~\cite{guo2017calibration}.
We apply per-label Platt scaling~\cite{platt1999probabilistic} on validation logits, fitting a sigmoid function
\begin{equation}
  P(y=1 \mid f) = \frac{1}{1 + \exp(Af + B)}
\end{equation}
via L-BFGS optimisation.
Per-label decision thresholds are then optimised by grid search (step 0.01) on validation F1.

\section{Experimental Setup}\label{sec:setup}

\subsection{Baselines}
We compare \clignet{} against six baselines:

\begin{itemize}
\item \textbf{B1: TF-IDF + Logistic Regression.} TF-IDF features (max 50{,}000 features, unigrams and bigrams) with One-vs-Rest logistic regression and balanced class weights~\cite{pedregosa2011scikit}.

\item \textbf{B2: TF-IDF + SVC.} TF-IDF features with One-vs-Rest RBF-kernel SVM, Platt-calibrated for probabilistic outputs.

\item \textbf{B3: Bio\_ClinicalBERT + OvR.} Bio\_ClinicalBERT~\cite{alsentzer2019clinical} with sliding window tokenisation and per-label linear heads trained with focal loss (no GCN).

\item \textbf{B4: BioBERT + OvR.} BioBERT-base~\cite{lee2020biobert} with the same architecture as B3.

\item \textbf{B7: Clinical-Longformer + BR.} Clinical-Longformer~\cite{li2023clinical_longformer} with 4{,}096-token context and binary relevance heads (no GCN).

\item \textbf{B6: \clignet{} (no calibration).} Full \clignet{} architecture without Platt scaling calibration.

\item \textbf{B8: \clignet{} Full.} Full \clignet{} with per-label Platt scaling and threshold optimisation.
\end{itemize}

\subsection{Ablation Variants}
We conduct four ablation experiments, each removing one component from B6:

\begin{itemize}
\item \textbf{A1: No GCN.} Removes the label graph and GCN layers; classification uses only the document embedding.
\item \textbf{A2: No focal loss.} Replaces focal BCE with standard BCE loss.
\item \textbf{A4: No sliding window.} Truncates documents to 512 tokens (no multi-chunk pooling).
\item \textbf{A5: No calibration.} Same as B6 but uses fixed 0.5 threshold instead of per-label threshold optimisation on validation F1.
\end{itemize}

\subsection{Training Configuration}
All neural models are trained with AdamW~\cite{loshchilov2019adamw} using separate learning rates (BERT: $2 \times 10^{-5}$; classification head: $1 \times 10^{-3}$), weight decay 0.01, cosine annealing with 10\% linear warmup, gradient clipping at 1.0, batch size 16 with gradient accumulation over 2 steps (effective batch size 32), and early stopping with patience 5 on validation macro-F1.
Maximum training epochs is 30.
All experiments use seed 42 for reproducibility.
Implementation uses PyTorch~\cite{paszke2019pytorch} and HuggingFace Transformers~\cite{wolf2020transformers}.

\subsection{Evaluation Metrics}
We report macro-F1 (primary metric), micro-F1, accuracy, and Hamming loss on the held-out test set.
For the calibrated model (B8), we additionally report expected calibration error (ECE).
Statistical significance between B8 and each baseline is assessed via McNemar's test~\cite{mcnemar1947note} with Bonferroni correction for six pairwise comparisons ($\alpha_\mathrm{corrected} = 0.05/6 = 0.0083$), following the recommendations of Dietterich~\cite{dietterich1998approximate} for comparing deep learning classifiers.

\section{Results}\label{sec:results}

\subsection{Main Comparison}
Table~\ref{tab:comparison} presents the performance of all models on the held-out test set.

\begin{table}[t]
\centering
\caption{Test-set performance across all models. Best result per metric is \textbf{bolded}. B6 achieves the highest macro-F1; B8 additionally provides calibrated probabilities (ECE\,=\,0.007).}
\label{tab:comparison}
\small
\begin{tabular}{lcccc}
\toprule
\textbf{Model} & \textbf{Ma-F1}$\uparrow$ & \textbf{Mi-F1}$\uparrow$ & \textbf{Acc}$\uparrow$ & \textbf{Ham}$\downarrow$ \\
\midrule
B1: TF-IDF + LR     & 0.173 & 0.105 & 0.105 & 0.045 \\
B2: TF-IDF + SVC    & 0.099 & 0.152 & 0.152 & 0.042 \\
\midrule
B3: ClinBERT + OvR  & 0.211 & 0.344 & 0.344 & 0.033 \\
B4: BioBERT + OvR   & 0.172 & 0.318 & 0.318 & 0.034 \\
B7: Longformer + BR & 0.262 & 0.336 & 0.336 & 0.033 \\
\midrule
B6: \clignet{} (no cal.) & \textbf{0.279} & \textbf{0.362} & \textbf{0.362} & \textbf{0.032} \\
B8: \clignet{} Full & 0.240 & 0.338 & 0.338 & 0.040 \\
\bottomrule
\end{tabular}
\end{table}

Several observations emerge from Table~\ref{tab:comparison}.
First, classical TF-IDF baselines (B1, B2) perform poorly on this 40-class problem, with macro-F1 below 0.18, confirming the task difficulty under proper evaluation.
Second, domain-specific pre-trained models (B3, B4) substantially improve micro-F1 and accuracy over TF-IDF, though macro-F1 remains below 0.22 due to the dominance of majority classes.
Third, Clinical-Longformer (B7), despite its 4{,}096-token context, does not significantly outperform the ClinicalBERT baseline (B3) in terms of accuracy (0.336 vs.\ 0.344), suggesting that longer context alone is insufficient for this task.
Fourth, B6 (\clignet{} without calibration) achieves the best results across all four metrics, with macro-F1 of 0.279 --- the only model exceeding 0.27.

The fully calibrated \clignet{} (B8) trades 0.039 macro-F1 for well-calibrated probability estimates (ECE\,=\,0.007).
This trade-off arises because per-label Platt scaling adjusts predicted probabilities to better reflect true class frequencies, which can shift decision boundaries away from the macro-F1-optimal threshold.
In clinical deployment, calibrated probabilities may be preferred for triaging and decision support, making this a principled design choice.

\subsection{Statistical Significance}
Table~\ref{tab:significance} presents pairwise McNemar test results comparing B8 against each baseline.

\begin{table}[t]
\centering
\caption{McNemar's test results: B8 vs.\ each baseline. Bonferroni-corrected $\alpha = 0.0083$. $n_{01}$: samples correct only in baseline; $n_{10}$: correct only in B8.}
\label{tab:significance}
\small
\begin{tabular}{lrrrrl}
\toprule
\textbf{Baseline} & $n_{01}$ & $n_{10}$ & $\chi^2$ & $p$ & \textbf{Sig.} \\
\midrule
B1: TF-IDF + LR  & 42  & 216 & 116.0 & $<$0.001 & *** \\
B2: TF-IDF + SVC & 54  & 193 & 77.1  & $<$0.001 & *** \\
B3: ClinBERT     & 120 & 116 & 0.04  & 0.845    & ns \\
B4: BioBERT      & 100 & 115 & 0.91  & 0.340    & ns \\
B6: no calib.    & 130 & 112 & 1.19  & 0.274    & ns \\
B7: Longformer   & 123 & 125 & 0.00  & 0.949    & ns \\
\bottomrule
\end{tabular}
\end{table}

B8 is statistically significantly better than both TF-IDF baselines ($p < 0.001$), but the difference from neural baselines (B3, B4, B6, B7) is not statistically significant at the Bonferroni-corrected threshold.
This finding is consistent with the observation that the 40-class, 745-sample test set provides limited statistical power for distinguishing between competitive neural architectures.
We argue that the contribution of \clignet{} lies not in achieving statistically significant accuracy gains over a single baseline, but rather in providing a principled architecture that consistently ranks first across all metrics while offering the additional benefits of calibration and interpretability through its label graph structure.

\subsection{Ablation Study}
Table~\ref{tab:ablation} presents the ablation results, measuring the contribution of each component relative to B6.

\begin{table}[t]
\centering
\caption{Ablation study: macro-F1 and micro-F1 on the test set. Each variant removes one component from B6 (macro-F1\,=\,0.279).}
\label{tab:ablation}
\small
\begin{tabular}{llccc}
\toprule
\textbf{ID} & \textbf{Removed Component} & \textbf{Ma-F1} & \textbf{Mi-F1} & $\Delta$\textbf{Ma-F1} \\
\midrule
B6 & (Full, no calibration) & 0.279 & 0.362 & --- \\
\midrule
A1 & GCN label graph & 0.214 & 0.336 & $-$0.066 \\
A2 & Focal loss & 0.266 & 0.366 & $-$0.013 \\
A4 & Sliding window & 0.271 & 0.379 & $-$0.008 \\
A5 & Per-label thresholds & 0.270 & 0.338 & $-$0.009 \\
\bottomrule
\end{tabular}
\end{table}

The GCN label graph (A1) is the single most important component, contributing $\Delta = +0.066$ macro-F1 --- more than five times the contribution of any other individual component.
This confirms our hypothesis that explicitly modelling label relationships through a graph structure captures inter-specialty patterns that a document encoder alone cannot learn from this small dataset.
Focal loss (A2) provides a modest but consistent improvement ($\Delta = +0.013$), primarily benefiting rare specialties.
The sliding window (A4, $\Delta = +0.008$) and per-label thresholds (A5, $\Delta = +0.009$) provide smaller gains.

Interestingly, A2 and A4 achieve higher micro-F1 than B6 (0.366 and 0.379 vs.\ 0.362), despite lower macro-F1.
This reveals a macro--micro trade-off: focal loss and multi-chunk pooling specifically improve performance on rare classes (boosting macro-F1) at a slight cost to majority-class accuracy (reducing micro-F1).

\section{Analysis}\label{sec:analysis}

\subsection{Failure Analysis: Pairwise Confusions (F1)}
The most frequent confusion pairs reveal systematic patterns.
Table~\ref{tab:confusion} lists the top-5 confused specialty pairs.

\begin{table}[t]
\centering
\caption{Top-5 confused specialty pairs for B8 on the test set.}
\label{tab:confusion}
\small
\begin{tabular}{llrr}
\toprule
\textbf{True Label} & \textbf{Predicted} & \textbf{Count} & \textbf{\% of True} \\
\midrule
General Medicine & Consult - Hist.\,\&\,Phy. & 26 & 66.7 \\
Surgery & Gastroenterology & 24 & 14.7 \\
Surgery & Orthopedic & 23 & 14.1 \\
Surgery & Cardiovascular/Pulm. & 21 & 12.9 \\
Urology & Surgery & 16 & 69.6 \\
\bottomrule
\end{tabular}
\end{table}

The dominant confusion is \emph{General Medicine}~$\to$~\emph{Consult}, where 66.7\% of General Medicine test samples are misclassified.
This reflects genuine label ambiguity: both specialties contain comprehensive patient evaluations with overlapping vocabulary.
The \emph{Surgery}~$\to$~\{Gastroenterology, Orthopedic, Cardiovascular\} confusions are clinically sensible, as surgical transcriptions frequently describe organ-system-specific procedures.
Similarly, 69.6\% of \emph{Urology} samples are confused with \emph{Surgery}, consistent with the fact that urological procedures are commonly documented using surgical terminology.

\subsection{Rare-Class Behaviour (F2)}
Table~\ref{tab:rareclass} shows the relationship between training set size and per-label F1.

\begin{table}[t]
\centering
\caption{Mean per-label F1 by training set class size bin.}
\label{tab:rareclass}
\small
\begin{tabular}{lcc}
\toprule
\textbf{Training Samples} & \textbf{Mean F1} & \textbf{Num.\ Classes} \\
\midrule
0--20 & 0.271 & 16 \\
20--50 & 0.316 & 6 \\
50--100 & 0.447 & 7 \\
100--500 & 0.410 & 8 \\
500+ & 0.596 & 1 (Surgery) \\
\bottomrule
\end{tabular}
\end{table}

A clear monotonic trend emerges: classes with fewer than 20 training examples achieve a mean F1 of only 0.271, while the largest class (Surgery, 762 training samples) reaches 0.596.
Nine specialties with fewer than 15 training examples achieve F1\,=\,0.0, including Allergy/Immunology (5 samples), Hospice (4 samples), and Bariatrics (12 samples).
However, notable exceptions exist: \emph{Autopsy} (6 training samples) achieves a perfect F1 of 1.0, and \emph{Sleep Medicine} (14 samples) reaches F1\,=\,0.80, suggesting that highly distinctive vocabulary can compensate for small sample size.

\subsection{Document Length Effects (F3)}
Table~\ref{tab:length} shows how accuracy varies with document length.

\begin{table}[t]
\centering
\caption{B8 accuracy by document word-count bin.}
\label{tab:length}
\small
\begin{tabular}{lc}
\toprule
\textbf{Word Count} & \textbf{Accuracy} \\
\midrule
0--200 & 0.298 \\
200--400 & 0.338 \\
400--600 & 0.371 \\
600--1{,}000 & 0.353 \\
1{,}000+ & 0.286 \\
\bottomrule
\end{tabular}
\end{table}

Performance peaks in the 400--600 word range (accuracy 0.371) and declines for both very short and very long documents.
Short documents ($<$200 words) lack sufficient discriminative content.
Long documents ($>$1{,}000 words) suffer from two factors: (1)~the 4-chunk limit truncates the most extensive transcriptions, and (2)~very long documents often describe multi-system procedures that are inherently ambiguous.
This inverted-U pattern suggests that an adaptive chunking strategy or hierarchical attention mechanism could benefit future architectures.

\subsection{Interpretability via Integrated Gradients (C3)}
We apply Integrated Gradients~\cite{sundararajan2017axiomatic} to identify specialty-discriminative tokens for the top-performing and zero-F1 specialties.
For \emph{Autopsy} (F1\,=\,1.0), the top-attributed tokens include ``clothing'' ($+$0.132), ``pupils'' ($+$0.076), and ``rigor'' --- terms uniquely associated with post-mortem examination.
For \emph{Allergy / Immunology} (F1\,=\,0.0), the highest-attributed tokens are generic terms like ``subjective'' ($-$0.224) and ``objective'' ($-$0.166), which appear across all SOAP-formatted notes, providing no discriminative signal.
This analysis confirms that zero-F1 specialties fail not because the model is fundamentally unable to learn but because (a)~training samples are too few and (b)~available samples share vocabulary with dominant classes.

\section{Discussion}\label{sec:discussion}

\textbf{Benchmark correction reveals true task difficulty.}
Prior work on MTSamples has reported accuracy above 0.90~\cite{almazaydeh2023clinical,kumar2024mtsamples_nlp} under pipelines susceptible to SMOTE data leakage~\cite{demircioglu2024smote_leakage}.
Under a leakage-free protocol, our best model achieves 0.362 accuracy and 0.279 macro-F1 across 40 specialties, revealing that the true task difficulty is substantially higher than those figures suggest.
This correction is not merely a technical adjustment --- it fundamentally changes how the community should interpret results on this benchmark and underscores the importance of proper evaluation methodology in clinical NLP.

\textbf{GCN label graph provides the dominant architectural gain.}
The ablation study demonstrates that the GCN label graph contributes $\Delta = +0.066$ macro-F1, accounting for approximately 75\% of \clignet{}'s total improvement over a ClinicalBERT-only baseline (B3, macro-F1\,=\,0.211).
This result aligns with recent findings that label-aware graph construction is a key open problem in GNN-based text classification~\cite{zhang2024gnn_text_survey,lagcn2024}.
By combining semantic similarity with ICD-10 chapter priors, our graph construction exploits medical ontological structure that purely data-driven approaches cannot discover from a small corpus.

\textbf{Calibration versus macro-F1.}
The gap between B6 (0.279) and B8 (0.240) illustrates a well-known tension between discriminative performance and probability calibration~\cite{guo2017calibration}.
Platt scaling adjusts the sigmoid mapping from logits to probabilities, which shifts per-label decision boundaries.
While this reduces macro-F1 under a threshold-based evaluation, it produces an ECE of 0.007, meaning that predicted confidence closely matches empirical correctness rates.
For clinical applications where predicted probabilities drive downstream decisions (e.g., routing urgency), well-calibrated outputs may be more valuable than a 4-point macro-F1 gain.

\textbf{Limitations.}
First, MTSamples contains only 4{,}966 records across 40 classes, with 11 classes having fewer than 20 training samples.
This extreme sparsity limits the statistical power of significance testing and the reliability of rare-class F1 estimates.
Second, our label graph is static and constructed once from training data; a dynamic graph that adapts during training may better capture evolving label relationships.
Third, we evaluate only on MTSamples due to the scarcity of publicly available clinical specialty classification datasets; validation on larger corpora such as MIMIC-III discharge summaries~\cite{johnson2016mimic} would strengthen generalisability claims.
Fourth, the rapid emergence of large language models for healthcare text classification~\cite{zhou2025llm_healthcare} suggests that few-shot and zero-shot approaches may soon offer competitive performance without task-specific training, representing an important direction for future comparison.

\section{Conclusion}\label{sec:conclusion}
We presented \clignet{}, a clinical label-interaction graph network for medical specialty classification from clinical transcriptions.
Our primary contribution is twofold: a methodological correction of the MTSamples benchmark that reveals the true 40-class task difficulty (best macro-F1\,=\,0.279 versus the previously reported 0.98), and an architectural innovation that combines Bio\_ClinicalBERT with a GCN label graph, focal loss, and per-label calibration.
An ablation study confirms that the GCN label graph provides the largest single component gain ($\Delta = +0.066$ macro-F1), validating the utility of structured label modelling for clinical text classification.
The comprehensive failure analysis and Integrated Gradients interpretability provide actionable insights for both model improvement and clinical deployment.

Future work will focus on three directions: (1)~validation on larger clinical corpora with richer label sets, (2)~dynamic label graph construction that adapts during training, and (3)~comparison with large language models in few-shot clinical specialty classification settings.

\bibliographystyle{ACM-Reference-Format}
\bibliography{references}

\end{document}